\documentclass[sigconf]{acmart}
\usepackage{multirow}
\usepackage{pifont}
\AtBeginDocument{%
  }

\copyrightyear{2025}
\acmYear{2025}
\setcopyright{acmlicensed}
\acmConference[MM '25] {Proceedings of the 33rd ACM International Conference on Multimedia}{October 27--31, 2025}{Dublin, Ireland.}
\acmBooktitle{Proceedings of the 33rd ACM International Conference on Multimedia (MM '25), October 27--31, 2025, Dublin, Ireland}
\acmISBN{979-8-4007-2035-2/2025/10}
\acmDOI{10.1145/3746027.3754806}

\begin{document}

\title[UniRGB-IR: A Unified Framework for Visible-Infrared Semantic Tasks via Adapter Tuning]{UniRGB-IR: A Unified Framework for Visible-Infrared \\ Semantic Tasks via Adapter Tuning}

\author{Maoxun Yuan}
\affiliation{%
  \institution{Institute of Artificial Intelligence, Beihang University}
  \city{Beijing}
  \country{China}
}\email{mxyuan@buaa.edu.cn}

\author{Bo Cui}
\affiliation{%
  \institution{School of Computer Science and Engineering, Beihang University}
  \city{Beijing}
  \country{China}
}  \email{tsuipo@outlook.com}

\author{Tianyi Zhao}
\affiliation{%
  \institution{Institute of Artificial Intelligence, Beihang University}
  \city{Beijing}
  \country{China}
}\email{ty_zhao@buaa.edu.cn}

\author{Jiayi Wang}
\affiliation{%
  \institution{CTTL-Terminal, China Academy of Information and Communications Technology}
  \city{Beijing}
  \country{China}
}\email{wangjiayi@caict.ac.cn}

\author{Shan Fu}
\affiliation{%
  \institution{CTTL-Terminal, China Academy of Information and Communications Technology}
  \city{Beijing}
  \country{China}
} \email{fushan@caict.ac.cn}

\author{Xue Yang}
\affiliation{%
  \institution{School of Automation and Intelligent Sensing, Shanghai Jiao Tong University}
  \city{Shanghai}
  \country{China}
}\email{yangxue-2019-sjtu@sjtu.edu.cn}

\author{Xingxing Wei}
\authornotemark[2]
\affiliation{%
  \institution{Institute of Artificial Intelligence, Beihang University}
  \city{Beijing}
  \country{China}
}\email{xxwei@buaa.edu.cn}
\thanks{$^\dagger$ Corresponding Author.}

\renewcommand{\shortauthors}{Maoxun Yuan et al.}

\begin{abstract}
Semantic analysis on visible (RGB) and infrared (IR) images has gained significant attention due to their enhanced accuracy and robustness under challenging conditions including low-illumination and adverse weather. However, due to the lack of pre-trained foundation models on the large-scale infrared image datasets, existing methods prefer to design task-specific frameworks and directly fine-tune them with pre-trained foundation models on their RGB-IR semantic relevance datasets, which results in poor scalability and limited generalization. To address these limitations, we propose UniRGB-IR, a scalable and efficient framework for RGB-IR semantic tasks that introduces a novel adapter mechanism to effectively incorporate rich multi-modal features into pre-trained RGB-based foundation models. Our framework comprises three key components: a vision transformer (ViT) foundation model, a Multi-modal Feature Pool (MFP) module, and a Supplementary Feature Injector (SFI) module. The MFP and SFI modules cooperate with each other as an adpater to effectively complement the ViT features with the contextual multi-scale features. During training process, we freeze the entire foundation model to inherit prior knowledge and only optimize the MFP and SFI modules. Furthermore, to verify the effectiveness of our framework, we utilize the ViT-Base as the pre-trained foundation model to perform extensive experiments. Experimental results on various RGB-IR semantic tasks demonstrate that our method can achieve state-of-the-art performance. The codes are available at \href{https://github.com/PoTsui99/UniRGB-IR}{https://github.com/PoTsui99/UniRGB-IR}.
\end{abstract}



\maketitle

\section{Introduction}

Image semantic analysis in the visible images is a common practice in computer vision and has been widely used in a variety of vision tasks such as object detection \cite{yan2022antijamming,bo2021ship}, instance segmentation \cite{khan2020survey,zhu2024parameter}, and semantic segmentation \cite{wang2025parameter}. 
However, visible cameras have demonstrated significant limitations in providing reliable imaging \cite{xu2017learning,yuan2024c2former}, due to their restricted spectral bandwidth. This constraint becomes particularly problematic under low-illumination conditions and adverse weather scenarios. Therefore, infrared (IR) imaging, with its superior low-light adaptability, has been increasingly employed as complementary information to enhance visible modality performance \cite{xu2017learning, zhao2024removal}. 
Afterwards, the joint use of RGB and IR images has been applied in more and more semantic analysis tasks.

With the rapid advancements in physical computing capabilities, more and more general-purpose foundation backbones \cite{dosovitskiy2020image, xia2024vit} pre-trained on large-scale RGB-based datasets (\emph{e.g.}, ImageNet \cite{deng2009imagenet} and COCO \cite{lin2014microsoft}) have been designed for various tasks. Since these pre-trained models implicitly encode substantial prior knowledge, it is believed that pre-trained models can promote the improvement of downstream task performance and speed up training convergence. Therefore, fine-tuning the pre-trained foundation models \cite{wei2023unified,shi2023open} on their semantic relevance datasets has become a common paradigm to improve the performance on downstream tasks.

\begin{figure}[!t]
    \centering
    \includegraphics[width=0.9\linewidth]{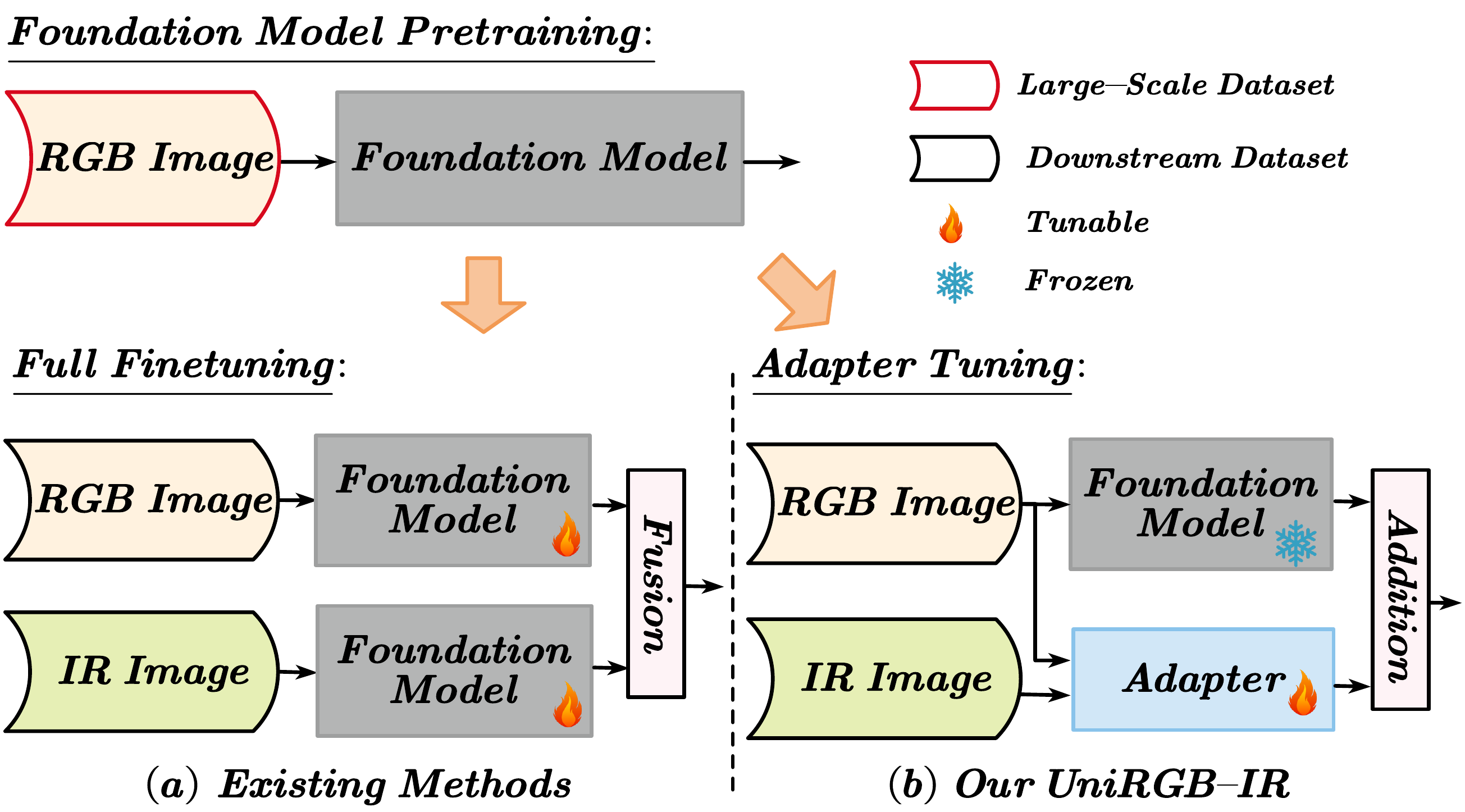}
    \caption{Existing full fine-tuning methods \emph{vs.} our UniRGB-IR framework. (a) Existing methods use pre-trained RGB-based foundation models and fully fine-tune them on their RGB-IR semantic relevance datasets. (b) We utilize the Adapter \cite{houlsby2019parameter} to propose a unified framework, which can efficiently introduce richer RGB-IR features into the pre-trained foundation model for various semantic tasks.}
    \label{fig:intro}
\end{figure}

However, due to the lack of pre-trained foundation models on the large-scale infrared image datasets, for RGB-Infrared semantic tasks, a straightforward and prevailing way \cite{zhang2021weakly,liu2021cross} is to use pre-trained RGB-based models and fine-tune them on their RGB-IR semantic relevance datasets as shown in Figure~\ref{fig:intro} (a). For example, RGB-IR object detectors \cite{zhang2021weakly,yuan2022translation} utilize RGB-based models as strong baselines and fine-tune them to extract RGB and IR features. SwinNet \cite{liu2021swinnet} employs transformer architecture to extract hierarchical features across RGB and infrared modalities, effectively detecting salient RGB-IR objects. Similarly, Ha \emph{et al.} \cite{ha2017mfnet} establish a novel baseline by incorporating infrared modality features into an RGB-based framework and fine-tune it on the RGB-IR semantic segmentation benchmark. Among the efforts to achieve competitive results, these approaches still suffer from two major issues:

\noindent\ding{202} \textbf{Limited model versatility and redundant parameter storage.} Due to the lack of a unified framework for RGB-IR semantic tasks, different semantic tasks often require customized model structures tailored to specific objectives. However, task-specific customization often leads to poor model versatility, as each task demands its own dedicated model structure. Besides, maintaining multiple specialized models for different downstream tasks results in excessive parameter storage requirements.

\noindent\ding{203} \textbf{Impairment of prior knowledge encoded in foundation models.}
Pre-trained foundation models are typically initialized using large-scale datasets, where they learn rich feature representations that capture general visual patterns. However, the full fine-tuning strategy indiscriminately updates the model parameters to adapt to the task-specific RGB-IR datasets, which often overrides the prior knowledge encoded in the pre-trained model. This will reduce the generalization potential of the fine-tuned model.

\textit{\textbf{Above challenges necessitate the adaptation of RGB-based foundation models to construct a unified framework capable of handling RGB-IR semantic tasks effectively.}}

Drawing inspiration from recent advances in adapters \cite{houlsby2019parameter,stickland2019bert}, which are initially used to yield an extensible model to effectively exploit the potential representation of foundation models in the natural language processing (NLP) field, we develop an adapter to dynamically introduce extensive RGB-IR features into the pre-trained RGB-based foundation model. In our adapter, a feature extractor is designed to obtain rich RGB-IR features and a feature injector is proposed to adaptively introduce the required features for the foundation model. Without altering the original RGB-based foundation model, the robust pre-trained weights can be directly preserved to expedite training convergence and enable efficient fine-tuning for downstream tasks. Consequently, in this paper, we establish a \textbf{Uni}fied framework for \textbf{RGB-IR} semantic tasks, termed \textbf{UniRGB-IR}, as illustrated in Figure~\ref{fig:intro} (b).

\begin{figure*}[!t]
    \centering
    \includegraphics[width=0.85\linewidth]{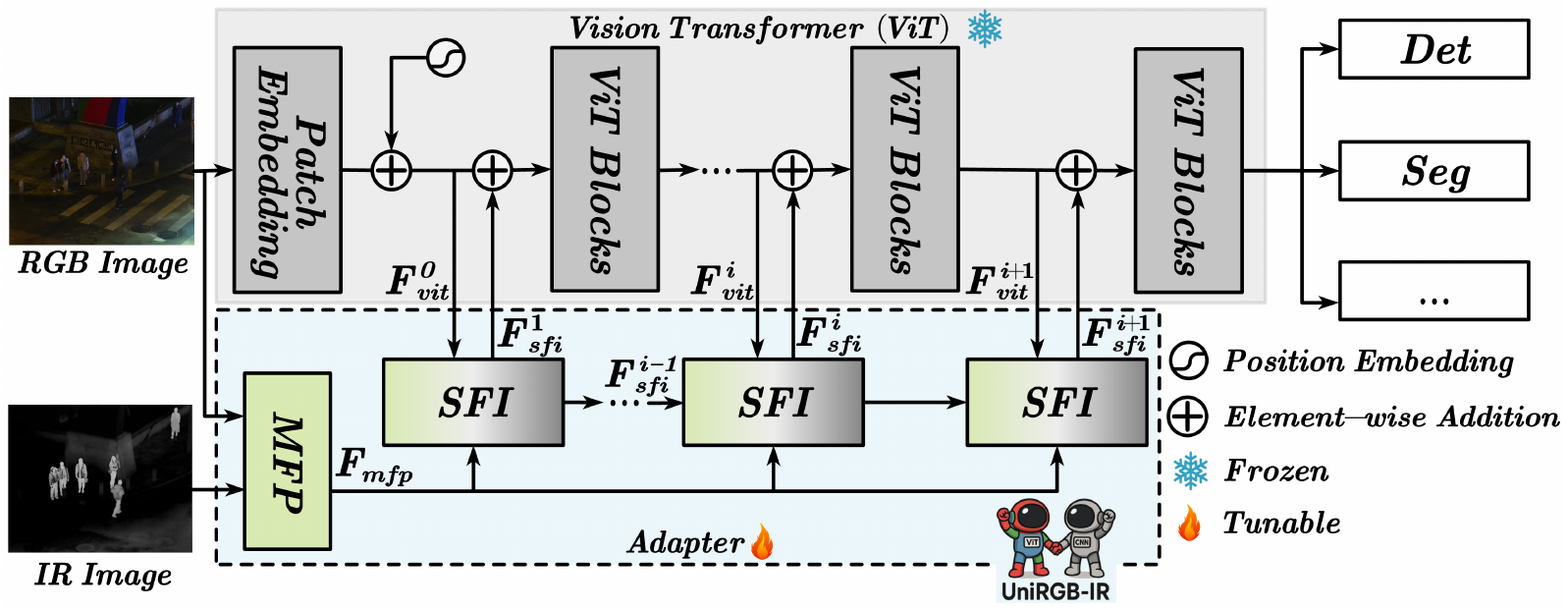}
    \caption{The overall architecture of our UniRGB-IR. In our framework, a ViT model with different numbers of ViT blocks is deployed as a foundation model, which is divided into $N$ (usually $N=4$) stages for feature interaction.
    During training, we freeze the entire ViT model weights and only optimize the MFP and SFI modules.
    }
    \label{fig:architecture}
\end{figure*}

Specifically, we design two core modules in our UniRGB-IR: a \textbf{M}ulti-modal \textbf{F}eature \textbf{P}ool module (MFP) and a \textbf{S}upplementary \textbf{F}eature \textbf{I}njector module (SFI). For the MFP module, the multi-receptive field convolution and the feature pyramid are deployed to capture contextual multi-scale features from RGB and IR images. These features are fused at different scales to complement foundation models for different RGB-IR semantic tasks. As for the SFI module, we implement a cross-attention mechanism that dynamically injects essential features into the pre-trained foundation model, endowing our UniRGB-IR framework with robust feature representation capabilities. To inherit prior knowledge of the foundation model pre-trained on the large-scale datasets, we utilize the adapter tuning paradigm instead of the full fine-tuning manner. We freeze the pre-trained weights and only optimize the adapter. Consequently, our UniRGB-IR can serve as a unified framework to achieve effective fine-tuning for various RGB-IR semantic tasks.

Overall, our contributions are summarized as follows:

\begin{itemize}
    \item We explore an scalable and efficient framework called UniRGB-IR for RGB-IR semantic tasks. To the best of our knowledge, this is the first attempt to construct a unified framework for various RGB-IR downstream tasks. 
    \item We design a Multi-modal Feature Pool module alongside a Supplementary Feature Injector module. The former extracts contextual multi-scale features from two modality images, and the latter dynamically injects the required features into the pre-trained model. These two modules can be efficiently fine-tuned with adapter tuning paradigm to complement the pre-trained foundation model with richer RGB-IR features for specific semantic task.
    \item We incorporate the vision transformer foundation model into the UniRGB-IR framework to evaluate the effectiveness of our method on RGB-IR semantic tasks, including RGB-IR object detection, RGB-IR semantic segmentation, and RGB-IR salient object detection. Extensive experimental results demonstrate that our methods can efficiently achieve superior performance on these downstream tasks.
\end{itemize}

\section{Related Work}

\subsection{Vision Foundation Models}

Recently, thanks to the powerful long-distance modeling capability, vision transformers (ViT) \cite{dosovitskiy2020image} have been widely used as the foundation model in many vision tasks to achieve competitive results. The original ViT is a plain, non-hierarchical structure for image classification. Based on it, ViTDet \cite{li2022exploring} also constructs a non-hierarchical model by incorporating the feature pyramid. However, the non-hierarchical structure lacks rich feature representation, resulting in unsatisfactory performance. Subsequently, various hierarchical transformers \cite{liu2021swin,fan2021multiscale,li2023mask,xia2024vit} have been proposed for different downstream vision tasks. Swin Transformer \cite{liu2021swin} and Multiscale Vision Transformer \cite{fan2021multiscale} are designed based on the ViT model to explore multi-scale features to improve the performance of image classification and object detection tasks. Besides, PVT \cite{wang2021pyramid} performs global attention on the downsampled key and value maps for dense prediction. In our work, we leverage the ViT model as our pre-trained foundation to build a unified framework for RGB-IR semantic tasks.

\subsection{RGB-IR Semantic Tasks}

\noindent\textbf{RGB-IR Object Detection.}
Zhang \emph{et al.} \cite{zhang2019cross} explore a two-stream SSD \cite{liu2016ssd} structure to capture the contextual enhanced features for RGB-IR object detection. Besides, AR-CNN \cite{zhang2019weakly} is presented based on Faster R-CNN \cite{ren2015faster} to align RGB and IR features. With the emergence of transformers, Yuan \emph{et al.} \cite{yuan2024improving} propose a complementary fusion transformer (CFT) module to achieve advanced detection results. Furthermore, C$^2$Former \cite{yuan2024c2former} is a novel transformer block and can be incorporated into exited pre-trained models to increase intra- and inter-modality feature representations.

\noindent\textbf{RGB-IR Semantic Segmentation.}
MFNet \cite{ha2017mfnet} is proposed to incorporate infrared features into the RGB-based framework to perform RGB-IR semantic segmentation. Based on the transformer structure, Wu \emph{et al.} \cite{wu2022complementarity} propose a novel CCFFNet to excavate discriminative and complementary modality features for RGB-IR semantic segmentation. Moreover, CMX \cite{zhang2023cmx} is designed as a universal cross-modal fusion framework for RGB-IR semantic segmentation in an interactive fusion manner.

\noindent\textbf{RGB-IR Salient Object Detection.}
SwinNet \cite{liu2021swinnet} is designed based on the Swin Transformer to extract hierarchical information of each modality, which achieves impressive results. CAVER \cite{pang2023caver} introduces the transformer to rethink the bi-modal salient object detection from a sequence-to-sequence perspective, which increases the model interpretability. Recently, Zhou \emph{et al.} \cite{zhou2023wavenet} transfer a large amount of knowledge learned in the transformer-based network to lightweight WaveNet through the distillation method.

The above methods attempt to design task-oriented structures to improve performance on corresponding downstream tasks. They either train the designed model from scratch or adopt a full fine-tuning strategy on a pre-trained model. Unlike these methods, we deploy an adapter based on the pre-trained foundation model for various RGB-IR semantic tasks. 

\begin{figure*}[!t]
    \centering
    \includegraphics[width=0.75\linewidth]{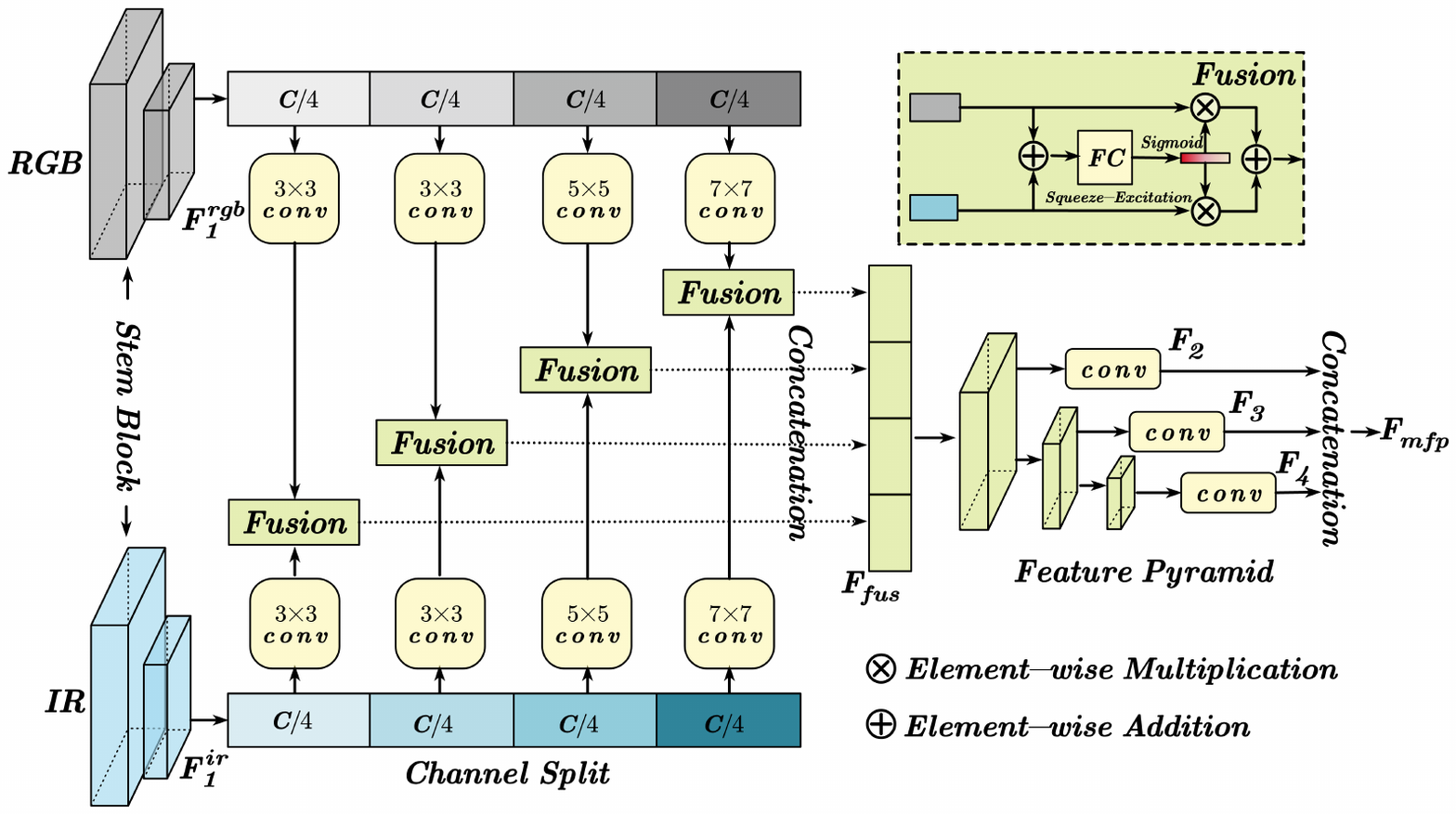}

    \caption{Structure of the multi-modal feature pool (MFP) module.  We explore multiple perceptions to expand the receptive field of contextual feature extraction and utilize the feature pyramid to obtain the multi-scale features. 
    }

    \label{fig:MFP}
\end{figure*}

\subsection{Adapters}
In the NLP field, Adapter \cite{houlsby2019parameter} first fixes the original foundation backbone and introduces new modules into the transformer for task-specific fine-tuning, thereby effectively adapting the pre-trained backbone to downstream NLP tasks. Afterwards, Adapter has been widely studied in computer vision. ViT-Adapter \cite{chen2022vision}, Low-Rank Adapter \cite{yin20231} and Mona \cite{yin20245} are designed to introduce a modest number of trainable parameters into the ViT and fine-tuning efficiently for dense prediction tasks. In addition, PC-Adapter \cite{park2023pc} explores an attention-based adapter to preserve global shape knowledge for domain adaptation on point cloud data. Recently, Adapter is also used as a parameter-efficient training technology for vision-and-language tasks \cite{sung2022vl,upadhyay2023probvlm}.

In this paper, we aim to explore an adapter capable of converting the IR features into the RGB features to fit the pre-trained foundation model, which remains a challenge to design this unified framework. Our UniRGB-IR is the first to propose utilizing adapter for RGB-IR semantic tasks. 

\section{Method}
\subsection{Overall Architecture}
The overall framework of UniRGB-IR is illustrated in Figure~\ref{fig:architecture}, which consists of three parts: vision transformer model, Multi-modal Feature Pool (MFP) module, and Supplementary Feature Injector (SFI) module. 
In our framework, the ViT model is utilized as the pre-trained foundation model and frozen during the training process. Specifically, for the ViT model, the RGB image is directly fed into the patch embedding process to obtain the $D$-dimensional feature tokens, which are usually 1/16 of the original image resolution. To complement the richer features required for various RGB-IR semantic tasks, we feed the RGB and IR images into the MFP module to extract contextual multi-scale features from two modalities (\emph{eg.} 1/8, 1/16 and 1/32 of the original image resolution). Afterwards, these richer features are dynamically injected into the features of ViT model through the SFI module, which can adaptively introduce the required RGB-IR features into the ViT model. To fully integrate the extracted features into the ViT model, we add the SFI module at the beginning of each stage. Consequently, after $N$ stages of feature injection, the final features from ViT model can be leveraged for various RGB-IR semantic tasks.

\begin{figure}[!t]
    \centering
    \includegraphics[width=0.75\linewidth]{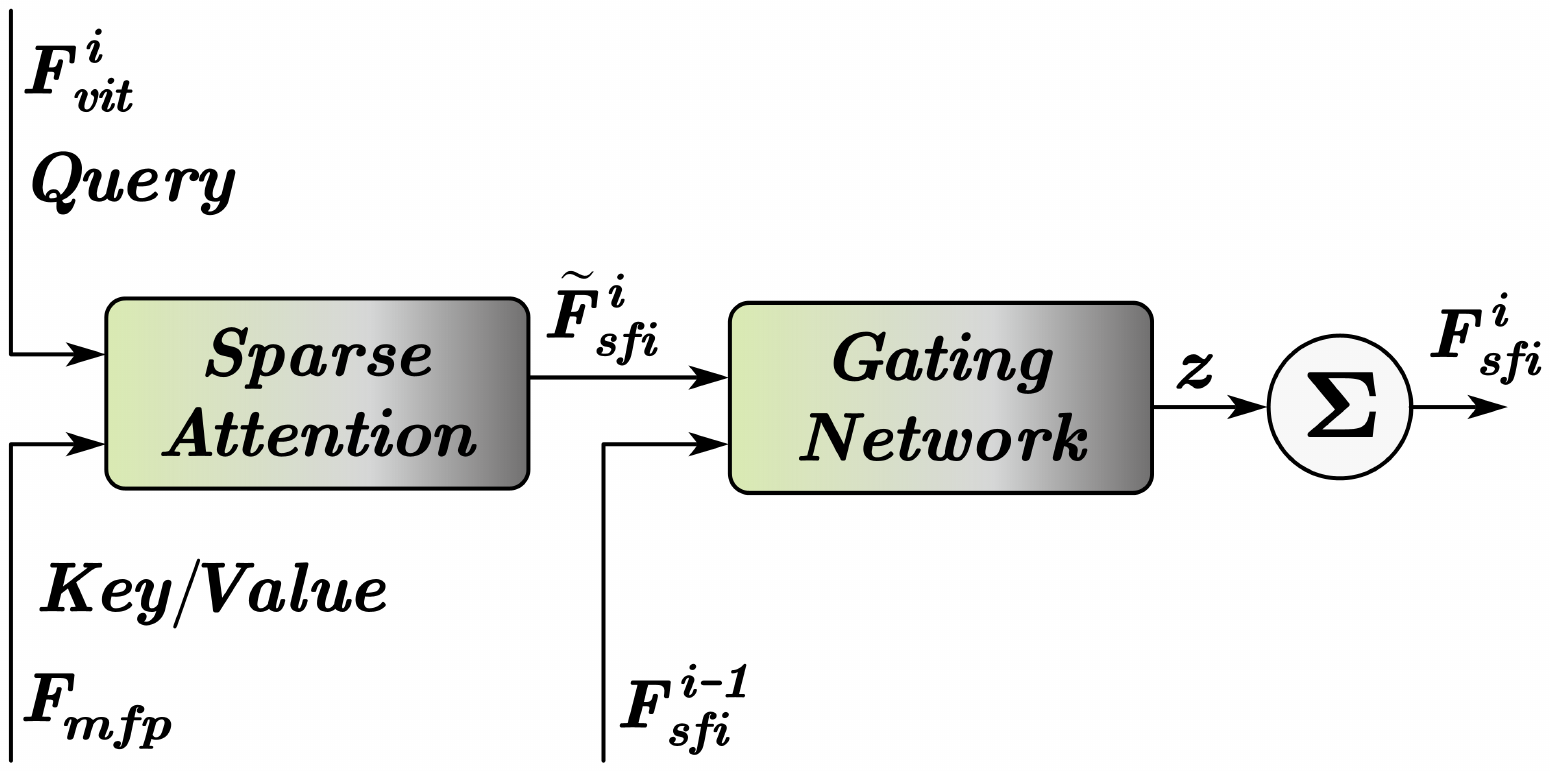}

    \caption{Structure of supplementary feature injector (SFI) module. A gating network is utilized to dynamically fuse the current and the last injected features.}

    \label{fig:SFI}
\end{figure}

\begin{table*}[]
 \caption{RGB-IR object detection results (mAP, in\%) on the FLIR and LLVIP dataset. The best results are highlighted in {\textcolor{red}{\textbf{red}}} and the second-best are highlighted in \textcolor{blue!70}{\textbf{blue}}. ``--'' indicates that the authors do not provide the corresponding results.}

 \vspace{-0.3cm}

    \label{tab:flir and llvip}
    \setlength{\tabcolsep}{3mm}
  \renewcommand{\arraystretch}{0.85}
\begin{tabular}{cc|ccc|ccc}
 \hline
& & \multicolumn{3}{c|}{\textbf{FLIR}}  & \multicolumn{3}{c}{\textbf{LLVIP}}   \\ \cline{3-8}
\multirow{-2}{*}{\textbf{Methods}}&\multirow{-2}{*}{\textbf{Modality}}& \textbf{mAP}& \textbf{$\text{mAP}_{50}$} & {\textbf{$\text{mAP}_{75}$}} & \textbf{mAP}& \textbf{$\text{mAP}_{50}$}& {\textbf{$\text{mAP}_{75}$}} \\\hline
SSD \cite{liu2016ssd}& IR & 24.6 & 57.0 & 18.0 & 53.5 & 90.2 &57.9  \\  
RetinaNet \cite{lin2017focal}& IR & 31.5 &66.1 &25.3   & 55.1&94.8 &57.6  \\
Faster R-CNN \cite{ren2015faster}& IR & 37.6 & 75.8 & 31.6 & 54.5 & 94.6 & 57.6 \\

Cascade R-CNN (ViT-B) \cite{cai2018cascade}& IR & 41.9 & 78.6 & 37.3  & 60.4 & 94.9 & 67.5 \\ 
DDQ-DETR \cite{zhang2023dense}& IR & 37.1 & 73.9   & 32.2   & 58.6 & 93.9 & 64.6 \\\hline
SSD \cite{liu2016ssd}& RGB & 18.8 & 46.3 & 13.1 & 39.8 & 82.6  &31.8  \\   
RetinaNet \cite{lin2017focal}&RGB & 21.9 &51.2 &15.2  & 42.8 &88.0& 34.4 \\ 
Faster R-CNN \cite{ren2015faster}&RGB & 27.7 & 62.2 & 21.2 & 45.1&87.0&41.2 \\ 
Cascade R-CNN (ViT-B) \cite{cai2018cascade}& RGB & 33.3 & 69.3 & 26.2 & 51.7 & 90.3 & 54.7 \\ 
DDQ-DETR \cite{zhang2023dense}& RGB & 30.9 & 64.9   &  24.5 & 46.7 & 86.1 &45.8  \\\hline

GAFF \cite{zhang2021guided} &RGB+IR & {37.4} & {74.6} & {31.3}  & {55.8} & {94.0} & {60.2} \\
ProbEn \cite{chen2022multimodal}     &RGB+IR & 37.9 & 75.5 & 31.8  & 51.5 & 93.4& 50.2 \\
CSAA \cite{cao2023multimodal} & RGB+IR & {41.3} & 79.2 & {37.4} & {59.2} & 94.3 & 66.6 \\ 
ICAFusion \cite{shen2024icafusion}  & RGB+IR& 41.4 & 79.2 & {36.9}  & {-} & {-}     & - \\
RSDet \cite{zhao2024removal} & RGB+IR & \textcolor{blue!70}{\textbf{43.8}} & \textcolor{red}{\textbf{83.9}} & \textcolor{blue!70}{\textbf {40.1}}  & \textcolor{blue!70}{\textbf{61.3}} & \textcolor{blue!70}{\textbf{95.8}} & \textcolor{blue!70}{\textbf{70.4}} \\ 
\hline
UniRGB-IR (Ours) &RGB+IR& \textcolor{red}{\textbf{44.1}} & \textcolor{blue!70}{\textbf{81.4}} &\textcolor{red}{\textbf {40.2}}  & \textcolor{red}{\textbf{63.2}} & \textcolor{red}{\textbf{96.1}}& \textcolor{red}{\textbf{72.2}} \\\hline
\end{tabular}
\end{table*}

\subsection{Multi-modal Feature Pool}
To complement the rich feature representations for RGB-IR semantic tasks, we introduce a multi-modal feature pool (MFP) module, including multiple perception and feature pyramid. The former can extract the contextual features with different convolution kernels, which achieve the long-distance modeling capability of CNNs. Different from existing works \cite{he2019mrfn,wu2022uiu} that increase the width or depth of the model, we efficiently achieve multi-receptive field perception in the channel dimension. As for feature pyramid, it can obtain multi-scale features to enhance the small object features. Therefore, these two operations are connected in series, enabling the MFP module to efficiently provide rich RGB-IR feature representations for various visible infrared semantic tasks, as shown in Figure~\ref{fig:MFP}.

Specifically, for the input RGB ($H\times W \times 3$) and IR ($H\times W$) images, we first employ a stem block borrowed from ResNet \cite{he2016deep} to extract two modality features $F^{rgb}_{1}$ and $F^{ir}_{1}\in \mathbb{R} ^{H/4\times W/4\times C}$. Then, these two features are split into four equal parts by utilizing channel splitting. To extract multi-receptive field perception, each part is subjected to convolution operations with different kernel sizes ($3\times3$, $3\times3$, $5\times5$ and $7\times7$). Then, we fuse each processed feature from two modalities by using SE attention \cite{hu2018squeeze} (shown in Figure~\ref{fig:MFP}). Therefore, we concatenate each fused part to obtain RGB-IR contextual features $F_{fus}$, which can be formulated as:
\begin{equation}
    \begin{aligned} 
F_{fus}  =\varGamma _{k=1}^{4}( Fus( W_k^{rgb}*f_{k}^{rgb},W_k^{ir}*f_{k}^{ir} ) ), 
    \end{aligned}
    \label{eq:MFP}
\end{equation}
where $F_{fus}\in \mathbb{R} ^{H/4\times W/4\times C}$, $f_{k}^{rgb}$ and $f_{k}^{ir}$ are the $k$-th part of $F_{1}^{rgb}$ and $F_{1}^{ir}$ features respectively, $W_k$ is the convolution with $k$-th kernel size, $\varGamma$ is the concatenation operation, $Fus(\cdot,\cdot)$ denotes the fusion module as shown in Figure~\ref{fig:MFP}.
For the feature pyramid, a stack of three $3 \times 3$ convolutions with $stride=2$ is applied to downsample the size of the feature maps. Then, features of each scale are fed into a $1\times1$ convolution to project the feature maps to $D$ dimensions. Therefore, we can obtain a set of multi-scale features \{$F_2$, $F_3$, $F_4$\} with 1/8, 1/16, and 1/32 of the original image resolution, respectively. Finally, we flatten and concatenate these features into feature tokens $F_{mfp}\in\mathbb{R}^{(\frac{H W}{8^2}+\frac{H W}{16^2}+\frac{H W}{32^2}) \times D}$, which will be used as supplementary features for the ViT foundation model.

\subsection{Supplementary Feature Injector}
To adaptively introduce the contextual multi-scale features without altering the ViT structure, we propose a supplementary feature injector (SFI) module as shown in Figure~\ref{fig:SFI}. Since the sequence lengths of contextual multi-scale features $F_{mfp}$ and ViT features $F_{vit}^i$ are different, to address this, we employ sparse attention (\emph{eg.} Pale Attention \cite{wu2022pale} and Deformable Attention \cite{zhu2020deformable}) to dynamically sample supplementary features from each scale. Specifically, we utilize the ViT features $F_{vit}^i \in \mathbb{R}^{\frac{H W}{16^2} \times D}$ as the query, and the contextual multi-scale features $F_{mfp}\in\mathbb{R}^{(\frac{H W}{8^2}+\frac{H W}{16^2}+\frac{H W}{32^2}) \times D}$ as the key and value, which can be represented as:
\begin{equation}
    \begin{aligned}
 \tilde{F}_{sfi}^{i}=Attention( LN( F_{vit}^{i} ), LN( F_{mfp} ) ) ,
    \end{aligned}
\end{equation}
where $Attention(\cdot)$ is the sparse attention and $LN(\cdot)$ is LayerNorm \cite{ba2016layer}, which aims to reduce modality differences during training. Furthermore, we adopt progressive injection to introduce contextual multi-scale features, which can balance the foundation model features and the injected features $F_{sfi}^i$. Thus, a gating network is explored to predict the fusion weight $z$ to gate $F_{sfi}^{i-1}$ and $\tilde{F}_{sfi}^{i}$ for dynamic fusion. Specifically, we concatenate the two features $F_{sfi}^{i-1}$ and $\tilde{F}_{sfi}^{i}$ and feed it into linear layer to predict the weight $z$. Then, $z$ and $1-z$ are used to fuse $F_{sfi}^{i-1}$ and $\tilde{F}_{sfi}^{i}$ features respectively. The final output features $F_{sfi}^{i}$ of SFI module can be formulated as:
\begin{equation}
    F_{sfi}^i= \begin{cases}\tilde{F}_{sfi}^i \ , & i=1 \\
    (1-z)*\tilde{F}_{sfi}^i + z * F_{sfi}^{i-1} \ . & i=2 \ldots N\end{cases}
\end{equation}

\subsection{Adapter Tuning Paradigm}
To fully inherit the prior knowledge of the ViT pre-trained on large-scale datasets, we explore the adapter tuning paradigm instead of the full fine-tuning manner. For the dataset $D=\{(x_j, {gt}_j)\}_{j=1}^M$ of the different semantic tasks, full fine-tuning process calculates the loss between the prediction and the ground truth, which can be formulated as:
\begin{equation}
    \mathcal{L} (D, \theta)=\sum_{j=1}^M \operatorname{loss}(F_\theta(x_j), {gt}_j),
\end{equation}
where loss represents the loss function and $F_\theta$ denotes the entire network parameterized by $\theta$. Afterwards, $\theta$ is optimized through the formula:
\begin{equation}
    \theta \leftarrow \underset{\theta}{\arg \min } \mathcal{L}(D, \theta) .
\end{equation}

However, in our adapter tuning paradigm, the parameter $\theta$ consists of two parts, one part is the parameter in the original ViT model $\theta_V$, and the other part is the parameter in our UniRGB-IR $\theta_A$. During training, we freeze the parameter $\theta_V$ and only optimize the parameter $\theta_A$. Thus, the loss function and optimization of our a can be represented as:
\begin{equation}
    \mathcal{L} (D, {\theta_V}, {\theta_A})=\sum_{j=1}^M \operatorname{loss}(F_{\theta_V, \theta_A}(x_j), {gt}_j),
\end{equation}

\begin{equation}
    \theta_A \leftarrow \underset{\theta_A}\arg \min \mathcal{L}(D, \theta_V, \theta_A).
\end{equation}

\begin{table*}[!tbp]
     \centering
    \caption{RGB-IR pedestrian detection results ($\text{MR}^{\text{-2}}$, in\%) under `All-dataset' settings of different pedestrian distances, occlusion levels, and light conditions (Day and Night) on the KAIST dataset. The best and second results are highlighted in {\textcolor{red}{\textbf{red}}} and \textcolor{blue!70}{\textbf{blue}}.}
    \label{tab:kaist}
    \setlength{\tabcolsep}{2.7mm}
    \renewcommand{\arraystretch}{0.85}
\begin{tabular}{c|cccccc|ccc}    \hline
Methods & Near  & Medium & Far   & None  & Partial & Heavy  & Day   & Night & All   \\    \hline
ACF~\cite{hwang2015multispectral}    & 28.74 & 53.67  & 88.20 & 62.94 & 81.40   & 88.08   & 64.31 & 75.06 & 67.74   \\
Halfway Fusion~\cite{liu2016multispectral} & 8.13  & 30.34  & 75.70 & 43.13 & 65.21   & 74.36   & 47.58 & 52.35 & 49.18   \\
FusionRPN+BF~\cite{konig2017fully} & 0.04  & 30.87  & 88.86 & 47.45 & 56.10   & 72.20   & 52.33 & 51.09 & 51.70   \\
IAF R-CNN~\cite{li2019illumination}  & 0.96  & 25.54  & 77.84 & 40.17 & 48.40   & 69.76   & 42.46 & 47.70 & 44.23   \\
IATDNN+IASS~\cite{guan2019fusion}    & \textcolor{blue!70}{\textbf{0.04}}  & 28.55  & 83.42 & 45.43 & 46.25   & 64.57   & 49.02 & 49.37 & 48.96   \\
CIAN~\cite{zhang2019cross} & 3.71  & 19.04  & 55.82 & 30.31 & 41.57   & 62.48   & 36.02 & 32.38 & 35.53   \\
MSDS-R-CNN~\cite{li2018multispectral} & 1.29  & 16.19  & 63.73 & 29.86 & 38.71   & 63.37   & 32.06 & 38.83 & 34.15   \\
AR-CNN~\cite{zhang2019weakly}  & {0.00}  & 16.08  & 69.00 & 31.40 & 38.63   & {\textcolor{blue!70}{\textbf{55.73}}}   & 34.36 & 36.12 & 34.95   \\
MBNet~\cite{zhou2020improving} & {0.00}  & 16.07  & 55.99 & 27.74 & 35.43   & 59.14   & 32.37 & 30.95 & 31.87   \\
TSFADet~\cite{yuan2022translation}    & {0.00}  & 15.99  & 50.71 & 25.63 & 37.29   & 65.67   & 31.76 & 27.44 & 30.74  \\ 
CMPD~\cite{li2022confidence}  & {0.00}   & \textcolor{red}{\textbf{12.99}}  & 51.22 & 24.04 & 33.88  &59.37   & \textcolor{blue!70}{\textbf{28.30}} &  30.56 & 28.98   \\
CAGTDet~\cite{yuan2024improving}& {0.00}  & 14.00  & 49.40 & 24.48 & 33.20   & 59.35   & 28.79 & 27.73 & 28.96   \\ 
C$^2$Former~\cite{yuan2024c2former}  & {0.00}  & 13.71  & \textcolor{blue!70}{\textbf{48.14}} & \textcolor{blue!70}{\textbf{23.91}} & \textcolor{blue!70}{\textbf{32.84}}   & 57.81   & 28.48 & \textcolor{blue!70}{\textbf{26.67}} & \textcolor{blue!70}{\textbf{28.39}}  \\ \hline
UniRGB-IR (Ours)  & {\textcolor{red}{\textbf{0.00}}}  & {\textcolor{blue!70}{\textbf{13.44}}} & {\textcolor{red}{\textbf{38.21}}} & {\textcolor{red}{\textbf{20.26}}} & \textcolor{red}{\textbf{31.67}}   & \textcolor{red}{\textbf{55.03}}   & {\textcolor{red}{\textbf{25.93}}} & {\textcolor{red}{\textbf{23.95}}} & {\textcolor{red}{\textbf{25.21}}}   \\ \hline
\end{tabular}

\end{table*} 

\begin{table*}[!htbp]
\centering
    \caption{RGB-IR semantic segmentation on the PST900 dataset. The best results are highlighted in {\textcolor{red}{\textbf{red}}} and the second-best are highlighted in \textcolor{blue!70}{\textbf{blue}}. ``--'' indicates that the authors do not provide the corresponding results.} 
    \label{tab:PST900}
    \setlength{\tabcolsep}{2.5mm}
    \renewcommand{\arraystretch}{0.85}
\begin{tabular}{c|cc|cc|cc|cc|cc|cc}
\hline
\multirow{2}{*}{\textbf{Methods}} & \multicolumn{2}{c|}{\textbf{Background}} & \multicolumn{2}{c|}{\textbf{Fire-Exting.}} & \multicolumn{2}{c|}{\textbf{Backpack}} & \multicolumn{2}{c|}{\textbf{Hand-Drill}} & \multicolumn{2}{c|}{\textbf{Survivor}} & \multirow{2}{*}{\textbf{mAcc}} & \multirow{2}{*}{\textbf{mIoU}} \\ 
& \textbf{Acc}       & \textbf{IoU}       & \textbf{Acc}& \textbf{IoU}          & \textbf{Acc}      & \textbf{IoU}      & \textbf{Acc}       & \textbf{IoU}       & \textbf{Acc}      & \textbf{IoU}      &&\\\hline
MFNet~\cite{ha2017mfnet}       & 99.9   & 98.7 & 71.8 & 67.4   &52.8 &52.4  & 46.7   & 39.3 & 18.8  & 18.9  & 58.0   & 55.3    \\
RTFNet~\cite{sun2019rtfnet}   & 99.9 	&99.0	&78.6	&54.8	&62.5	&60.8	&76.7	&61.0	&65.2	&62.5	&76.6	&67.6 \\
CCNet~\cite{huang2019ccnet} &99.6	&98.7	&88.1	&73.8	&76.0	&73.0	&54.1	&51.0	&49.5	&33.5	&73.46	&66.0\\

ACNet~\cite{hu2019acnet} &99.8	&99.3	&84.9	&60.0	&85.6	&83.2	&53.6	&51.5	&69.1	&65.2	&78.7	&71.8 \\
PSTNet~\cite{shivakumar2020pst900}  & -   & 98.9& - & 70.1   & -  & 69.2  & -   & 53.6& -  & 50.0 & -         & 68.4      \\
FDCNet~\cite{zhao2022feature}      & 99.7  & 99.2 & 91.8 & 71.5  & 77.5 & 72.2 & 82.5 & 70.4 & 78.4 & 72.4 & 86.0 & 77.1 \\
CCFFNet~\cite{wu2022complementarity}     & \textcolor{red}{\textbf{99.9}}  & 99.4 & 87.0 & \textcolor{blue!70}{\textbf{79.3}}  & 80.2 & 77.8 & 88.3 & \textcolor{blue!70}{\textbf{77.4}}  & 77.9 & 73.4 & 86.7 & 81.5 \\
RSFNet~\cite{li2023residual}    & -   & 99.4 & - & 75.4   & -  & 84.9  & -   & 72.9& -  & 70.1  & -         & 80.5      \\
MMDRNet~\cite{liang2023mask} & 99.5 & 98.9 &77.9 &52.4 &79.1 &71.1 &77.1 &40.6 &69.8 &62.3 &81.3 &68.7 \\
SGFNet~\cite{wang2023sgfnet} &99.8  &99.4   &89.4 &75.6  &90.4   & \textcolor{blue!70}{\textbf{85.4}}   &\textcolor{blue!70}{\textbf{94.0}}   &76.7   &82.7  &\textcolor{blue!70}{\textbf{76.7}}   &\textcolor{blue!70}{\textbf{91.2}}   &82.8\\
MDRNet~\cite{zhao2024mitigating} & 99.3 & 99.1 &90.2 &63.0 &\textcolor{blue!70}{\textbf{93.1}} &76.3 &86.6 &63.5 &\textcolor{blue!70}{\textbf{85.6}} &71.3 &91.0 &74.6 \\
KDSNet~\cite{zhou2024knowledge}&\textcolor{blue!70}{\textbf{99.8}}	&\textcolor{blue!70}{\textbf{99.4}}	&\textcolor{blue!70}{\textbf{92.0}}	&\textcolor{red}{\textbf{82.3}}	&83.2	&81.2	&86.1	&71.3	&72.6	&70.5	&86.8	&\textcolor{blue!70}{\textbf{80.9}} \\
\hline
UniRGB-IR (Ours)& 99.7  & \textcolor{red}{\textbf{99.5}}  & \textcolor{red}{\textbf{97.0}}   & 72.0     & \textcolor{red}{\textbf{93.3}} & \textcolor{red}{\textbf{87.7}} & \textcolor{red}{\textbf{95.5}} & \textcolor{red}{\textbf{78.0}}  & \textcolor{red}{\textbf{86.1}} & \textcolor{red}{\textbf{77.8}} & \textcolor{red}{\textbf{94.3}}     & \textcolor{red}{\textbf{82.8}}  \\
\hline
\end{tabular}
\end{table*}

\section{Experiments}
 To evaluate the effectiveness of our UniRGB-IR, we utilize the ViT-Base model (pre-trained on COCO \cite{lin2014microsoft} dataset) as the foundation model and utilize this framework to perform RGB-IR semantic tasks. During training, we freeze the ViT-Base model and only optimize the MFP and SFI modules. We evaluate and compare our method with various competitive models, including CNN-based and Transformer-based models. Besides, our evaluation spans various tasks, including RGB-IR object detection on FLIR \cite{zhang2020multispectral}, LLVIP \cite{jia2021llvip}, and KAIST \cite{hwang2015multispectral} datasets, RGB-IR semantic segmentation on PST900 \cite{shivakumar2020pst900} and MFNet \cite{ha2017mfnet} (\emph{see supplementary materials}) datasets and RGB-IR salient object detection on VT821 \cite{wang2018rgb}, VT1000 \cite{tu2019rgb} and VT5000 \cite{tu2022rgbt}. Furthermore, ablation experiments on the designed modules and qualitative experiments are also conducted to verify that the UniRGB-IR framework can be leveraged as a unified framework to efficiently introduce RGB-IR features into the foundation model to achieve superior performance.

\subsection{RGB-IR Object Detection}
\noindent\textbf{Datasets.}
Our object detection experiments are based on the three paired RGB and IR object detection datasets. FLIR \cite{zhang2020multispectral} is a paired visible and infrared object detection dataset, including daytime and night scenes, which has 4,129 aligned RGB-IR image pairs for training and 1,013 for testing. For LLVIP \cite{jia2021llvip} dataset, it contains 15,488 aligned RGB-IR image pairs, of which 12,025 images are used for training and 3,463 images for testing. As for KAIST \cite{hwang2015multispectral} dataset, it is a aligned multispectral pedestrian deteciton dataset, in which 8,963 and 2,252 image pairs are utilized for training and testing.

\noindent\textbf{Metrics.}
For FLIR and LLVIP datasets, we employ mean Average Precision (mAP) to evaluate the detection performance. As for KAIST dataset, we use log-average miss rate $\text{MR}$ over the false positive per image (FPPI) with the range of [$10^{-2}$, $10^0$] to evaluate the pedestrian detection performance.

\noindent\textbf{Settings.}
All the experiments are conducted with NVIDIA GeForce RTX 3090 GPUs. We implement our framework on the MMDetection library and use the Cascade R-CNN \cite{cai2018cascade} as the basic framework to perform RGB-IR object detection. The detector is trained with an initial learning rate of 2$\times$10$^{-4}$ for 48 epochs. The batch size is set to 16, and the AdamW \cite{loshchilov2017decoupled} optimizer is employed with a weight decay of 0.1. Horizontal flipping is also used for data augmentation. 

\begin{table*}[ht]
     \centering
    \setlength{\tabcolsep}{1.8mm}
    \caption{RGB-IR salient object detection on VT821, VT1000 and VT5000 datasets. * represents RGB-D SOD transformed into RGB-T SOD. The best results are highlighted in {\textcolor{red}{\textbf{red}}} and the second-best are highlighted in \textcolor{blue!70}{\textbf{blue}}. }
    \renewcommand{\arraystretch}{0.85}
\begin{tabular}{c|cccc|cccc|cccc}
\hline
\multirow{2}{*}{\textbf{Model}} & \multicolumn{4}{c|}{\textbf{VT821}}    & \multicolumn{4}{c|}{\textbf{VT1000}}   & \multicolumn{4}{c}{\textbf{VT5000}}    \\ \cline{2-13} 
                       & \textbf{$\emph{S}\uparrow$}     & \textbf{$\emph{adpE}\uparrow$}  & \textbf{$\emph{adpF}\uparrow$}  & \textbf{$\emph{MAE}\downarrow$}   & \textbf{$\emph{S}\uparrow$}     & \textbf{$\emph{adpE}\uparrow$}  & \textbf{$\emph{adpF}\uparrow$}  & \textbf{$\emph{MAE}\downarrow$}   & \textbf{$\emph{S}\uparrow$}     & \textbf{$\emph{adpE}\uparrow$}  & \textbf{$\emph{adpF}\uparrow$}  & \textbf{$\emph{MAE}\downarrow$}   \\ \hline
MMCI* \cite{chen2019multi} & 0.763 & 0.784 & 0.618 & 0.087 & 0.886 & 0.892 & 0.803 & 0.039 & 0.827 & 0.859 & 0.714 & 0.055 \\
TANet* \cite{chen2019three} & 0.818 & 0.852 & 0.717 & 0.052 & 0.902 & 0.912 & 0.838 & 0.030 & 0.847 & 0.883 & 0.754 & 0.047 \\
S2MA* \cite{liu2020learning} & 0.811 & 0.813 & 0.709 & 0.098 & 0.918 & 0.912 & 0.848 & 0.029 & 0.853 & 0.864 & 0.743 & 0.053 \\
JLDCF* \cite{fu2021siamese} & 0.839 & 0.830 & 0.726 & 0.076 & 0.912 & 0.899 & 0.829 & 0.030 & 0.861 & 0.860 & 0.739 & 0.050 \\
MTMR \cite{wang2018rgb}                  & 0.725 & 0.815 & 0.662 & 0.109 & 0.706 & 0.836 & 0.715 & 0.119 & 0.680 & 0.795 & 0.595 & 0.114 \\
M3S-NIR \cite{tu2019m3s}               & 0.723 & 0.859 & 0.734 & 0.140 & 0.726 & 0.827 & 0.717 & 0.145 & 0.652 & 0.780 & 0.575 & 0.168 \\
SGDL \cite{tu2019rgb}                  & 0.765 & 0.847 & 0.731 & 0.085 & 0.787 & 0.856 & 0.764 & 0.090 & 0.750 & 0.824 & 0.672 & 0.089 \\
FMSF \cite{zhang2019rgb}                  & 0.760 & 0.796 & 0.640 & 0.080 & 0.873 & 0.899 & 0.823 & 0.037 & 0.814 & 0.864 & 0.734 & 0.055 \\
MIDD \cite{tu2021multi}                  & 0.871 & 0.895 & 0.803 & 0.063 & 0.915 & 0.933 & 0.880 & 0.027 & 0.868 & 0.896 & 0.799 & 0.043 \\
ADF \cite{tu2022rgbt}                   & 0.810 & 0.842 & 0.717 & 0.077 & 0.910 & 0.921 & 0.847 & 0.034 & 0.864 & 0.891 & 0.778 & 0.048 \\
LSNet \cite{zhou2023lsnet}                  & \textcolor{blue!70}{\textbf{0.877}} & \textcolor{red}{\textbf{0.911}} & 0.827 & 0.070 &{0.924} & 0.936 & \textcolor{blue!70}{\textbf{0.887}} & 0.022 & 0.876 & 0.916 & 0.827 & 0.036 \\ 
UniTR \cite{guo2024unitr}
& 0.873 & 0.892 &\textcolor{red}{\textbf{ 0.827}} & \textcolor{red}{\textbf{ 0.033 }}
& \textcolor{blue!70}{\textbf{0.929}} & \textcolor{blue!70}{\textbf{0.941}} & 0.885 & \textcolor{blue!70}{\textbf{0.019}} 
&\textcolor{blue!70}{\textbf{0.883}}  & \textcolor{blue!70}{\textbf{0.926}} & \textcolor{blue!70}{\textbf{0.839}} & \textcolor{blue!70}{\textbf{0.032}} \\ 
\hline

UniRGB-IR (Ours)       & \textcolor{red}{\textbf{0.881}}  & \textcolor{blue!70}{\textbf{0.895}}  & \textcolor{blue!70}{\textbf{0.806}}  & \textcolor{blue!70}{\textbf{0.039}}  & \textcolor{red}{\textbf{0.939}}  & \textcolor{red}{\textbf{0.943}}  & \textcolor{red}{\textbf{0.894}}  & \textcolor{red}{\textbf{0.018}}  & \textcolor{red}{\textbf{0.906}}  & \textcolor{red}{\textbf{0.935}}  & \textcolor{red}{\textbf{0.849}}  & \textcolor{red}{\textbf{0.027}}  \\ \hline
\end{tabular}
\label{tab:SOD}
\end{table*}

\noindent\textbf{Results on FLIR and LLVIP datasets.}
We compare our method with five common mono-modality  methods and four competitive multi-modality methods. As shown in Table~\ref{tab:flir and llvip}, it can be seen that most of the multi-modality detectors are even worse than mono-modality detectors (\emph{eg.} Cascade R-CNN in IR modality). Since RGB features interfere with infrared feature under limited illumination conditions, it has a negative impact on the fused features utilized for object detection tasks. However, our UniRGB-IR can effectively solve this problem through the SFI module, enabling the detector to achieve better classification and localization processes. 

\noindent\textbf{Results on KAIST dataset.}
The quantitative results of the different methods on the KAIST dataset are shown in Table~\ref{tab:kaist}. The experiments are conducted under `All-dataset' settings~\cite{hwang2015multispectral}. We compare our UniRGB-IR with thirteen multi-modal object detection methods. Our model achieves the best performance on the `All', `Day', and `Night' conditions and the other four of five subsets~(`Near', `Far', `None', `Partial' and `Heavy'), and rank second in the `Medium' subset. Furthermore, our detector surpasses the previous best competitor C$^2$Former by 3.18\%  on the `All' condition, which indicates UniRGB-IR is robust to the complex scenes. 

\subsection{RGB-IR Semantic Segmentation}

\noindent\textbf{Datasets.}
Our semantic segmentation experiments are performed on the public RGB-IR semantic segmentation dataset PST900 \cite{shivakumar2020pst900}. The PST900 dataset is divided into 597 pairs for training and 288 pairs for testing, containing five categories (background, fire extinguisher, backpack, hand drill, and survivor). The dataset is divided into three parts: training, validation, and testing in a ratio of 2:1:1.

\noindent\textbf{Metrics.}
Two metrics are utilized to evaluate the performance of semantic segmentation, namely mean accuracy (mAcc) and mean intersection over union (mIoU). Both metrics are calculated by averaging the ratios of the intersection and union of all categories.

\begin{table}[!tbp]
\centering
    \caption{Ablation studies of key components on the FLIR and PST900 datasets. The best results are highlighted in bold.}
    \label{tab:components}
    \setlength{\tabcolsep}{1.8mm}
    \renewcommand{\arraystretch}{0.85}
\begin{tabular}{cc|ccc|cc}
\hline
    && \multicolumn{3}{c|}{\textbf{FLIR}}& \multicolumn{2}{c}{\textbf{PST900}}  \\ \cline{3-7}
\multirow{-2}{*}{\textbf{MFP}} & \multirow{-2}{*}{\textbf{SFI}} & \textbf{mAP}& \textbf{$\text{mAP}_{50}$} & {\textbf{$\text{mAP}_{75}$}}& \textbf{mAcc}& \textbf{mIoU}\\ \hline
   &  & 38.5 & 74.2 & 33.7 & 89.5 & 75.8\\  
\checkmark   &  & 41.2& 77.9& 36.5& 91.7& 78.8\\
\checkmark   & \checkmark & { \textbf{44.1}} & {\textbf{81.4}} & {\textbf{40.2}} & { \textbf{94.3}} & { \textbf{82.8}} \\\hline
\end{tabular}
\end{table}

\noindent\textbf{Settings.}
Similarly, as the RGB-IR object detection task, we incorporate our method into the SETR \cite{zheng2021rethinking} basic framework and implement it on the MMSegmentation library. The fine-tuning process spins a total of 10K iterations with an initial learning rate of 0.01. We employ the SGD optimizer and set the batch size to 16.

\begin{table}[!tbp]
\centering
    \caption{Ablation of adding SFI module to different stages.}
    \label{tab:stages}
    \setlength{\tabcolsep}{0.8mm}
    \renewcommand{\arraystretch}{0.85}
\begin{tabular}{cccc|ccc}
\hline
\multicolumn{4}{c|}{\textbf{Different Stages with SFI module}} & \multirow{2}{*}{\textbf{mAP}} & \multirow{2}{*}{\textbf{$\text{mAP}_{50}$}} & \multirow{2}{*}{\textbf{$\text{mAP}_{75}$}} \\ \cline{1-4}
\textbf{Stage 1} & \textbf{Stage 2} & \textbf{Stage 3} & \textbf{Stage 4} &       &          &       \\ \hline
 \checkmark  &     &     &    & 41.7        & 80.0          & 37.2        \\
 \checkmark  &   \checkmark     &    &    & {43.3} & { \textbf{81.6}} & {39.7} \\
 \checkmark  & \checkmark       & \checkmark       &  &{ \textbf{44.1}}        & 81.4        & {\textbf{40.2}}    \\
\checkmark   & \checkmark   & \checkmark   & \checkmark  & 42.7        & 79.9        & 38.2      \\ \hline
\end{tabular}
\end{table}

\begin{table}[!tbp]
\centering
    \caption{Ablation of the different attention mechanisms.}
    \label{tab:attention}
    \setlength{\tabcolsep}{0.6mm}
    \renewcommand{\arraystretch}{0.85}
\begin{tabular}{c|c|ccc}
\hline
\textbf{Attention Mechanism} & \textbf{Complexity} & \textbf{mAP}& \textbf{$\text{mAP}_{50}$} & {\textbf{$\text{mAP}_{75}$}}  \\ \hline
Global Attention \cite{vaswani2017attention} & Quadratic          & 36.1         & 72.8& 30.6\\
Pale Attention \cite{wu2022pale} & Linear & 31.6         & 68.4& 24.6\\
Deformable Attention \cite{zhu2020deformable}        & Linear & {\textbf{44.1}}         & {\textbf{81.4}}& {\textbf{40.2}} \\\hline      
\end{tabular}
\end{table}

\noindent\textbf{Results.}
The quantitative results of the different RGB-IR segmentation methods on the PST900 dataset are shown in Table~\ref{tab:PST900}. The comparison results show that our model significantly outperforms other methods. Specifically, our model obtains the best performance in terms of both the mACC and mIoU. Besides, our model performs competitive performance in the Backpack, Hand-Drill and Survivor, outperforming the second-best methods by 2.3\%, 0.6\% and 1.1\% IoU, which strongly demonstrates the effectiveness of our UniRGB-IR.

\subsection{RGB-IR Salient Object Detection}
\noindent\textbf{Datasets.}
Our salient object detection (SOD) experiments are performed on the three public datasets: VT821 \cite{wang2018rgb}, VT1000 \cite{tu2019rgb} and VT5000 \cite{tu2022rgbt}. 
The VT821 dataset includes 821 registered RGB and IR images. 
The VT1000 dataset contains 1000 registered RGB-IR images with simple scenes and aligned images. The VT5000 dataset is a recent large-scale RGB-IR dataset, including a full-day scene under various limited light conditions. As usual in \cite{tu2022rgbt}, we utilize 2500 image pairs in the VT5000 dataset as the training dataset, and the remaining image pairs along with the image pairs from the VT821 and VT1000 datasets are used as the test datasets.

\noindent\textbf{Metrics.}
Four metrics are utilized to evaluate the performance of salient object detection namely F-measure (\emph{adpF} $\uparrow$), E-Measure (\emph{adpE} $\uparrow$), S-Measure (\emph{S} $\uparrow$) and Mean absolute error (\emph{MAE} $\downarrow$). $\uparrow$ and $\downarrow$ denote the higher the better and the lower the better, respectively.

\noindent\textbf{Settings.}
As same as the RGB-IR semantic segmentation task, we incorporate our method into the SETR  basic framework and also implement it on the MMSegmentation library. The fine-tuning process spins a total of 10K iterations with an initial learning rate of 0.01. We use the SGD optimizer and set the batch size to 64. For convenience, all input images are resized to 224 $\times$ 224 for testing.

\noindent\textbf{Results.}
Table~\ref{tab:SOD} reports the quantitative comparison results. As can be seen from Table~\ref{tab:SOD}, our UniRGB-IR outperforms SOTA methods both on VT1000 and VT5000 datasets in all evaluation metrics. Specifically, the $S$, $adpE$, $adpF$ and $MAE$ matrics of our UniRGB-IR achieve 0.906, 0.935, 0.849 and 0.027 on VT5000, all of which are higher than the previous competitor UniT \cite {guo2024unitr}. These remarkable results clearly indicate that the saliency maps predicted by UniRGB-IR are close to the corresponding ground-truths.

\begin{figure}[!tbp]
    \centering
    \includegraphics[width=1\linewidth]{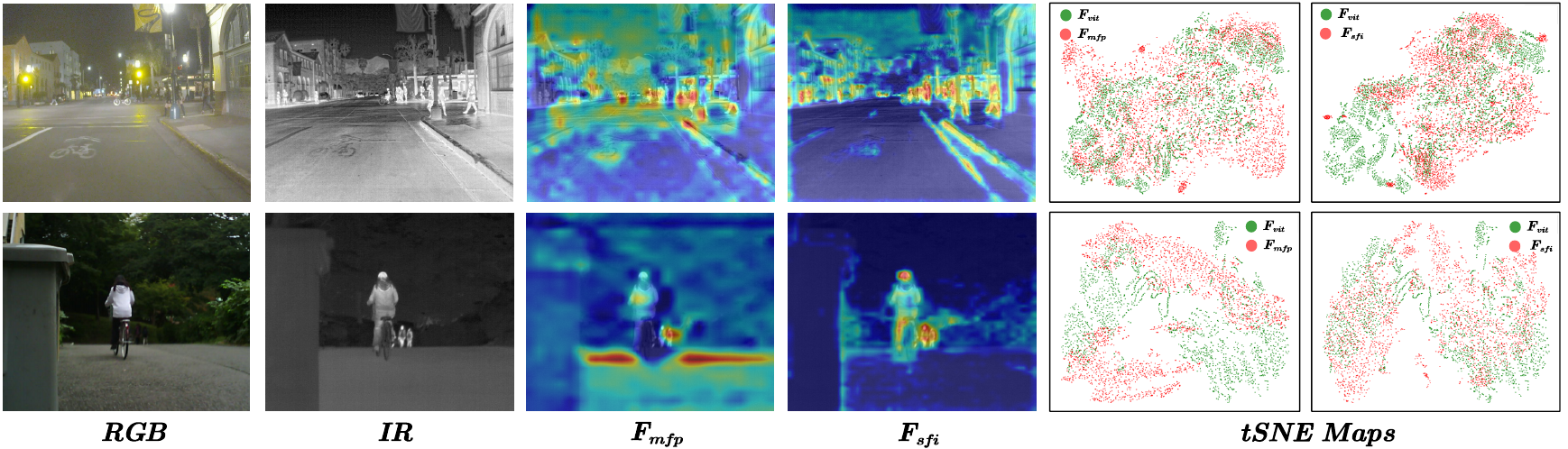}

    \caption{Visualization of intermediate results. The $\boldsymbol{F_{mfp}}$ and $\boldsymbol{F_{sfi}}$ features from the first stage are visualized in the third and fourth columns. The tSNE visualizations are also shown in the last two columns.}
    \label{fig:intermediate}
\end{figure}

\subsection{Ablation Study}
\noindent\textbf{Ablation for components.}
To investigate the contribution of the SFI and MFP modules, we gradually add each module to the baseline model as shown in Table~\ref{tab:components}. We stack RGB and IR images and feed them into a frozen standard ViT model equipped with the Cascade R-CNN detection head and SETR segmentation head. The results obtained by fine-tuning the patch embedding layer and the detection or segmentation head are considered as our baseline. Then, we also freeze the ViT model and introduce the MFP module to extract contextual multi-scale features (element-wise addition), which results in 2.7\% mAP and 3.0\% mIoU improvements. Finally, we replace the element-wise addition operation with the SFI module and further improve the mAP and mIoU metrics by 2.9\% and 4.0\% respectively, which achieve the best performance on both datasets.

\noindent\textbf{SFI module at different stages. }
Since we utilize the entire standard ViT encoder as the foundation model, we perform ablation experiments by adding the SFI module to the ViT pre-trained model at the beginning of different stages. From Table~\ref{tab:stages}, we can find that by adding the SFI module at the first stage, the detector achieves 41.7\% mAP on FLIR dataset. After adding the SFI module in the second and third stages, the performance further improved by about 2\% and 3\% mAP, respectively. However, continuing to add it to the final stage will reduce the detection performance while also increasing computational overhead. Therefore, we add the SFI module from the first stage to the third stage of ViT model.

\noindent\textbf{Attention type in SFI module. }
Since the attention mechanism in our SFI module is replaceable, we adopt three popular attention mechanisms in our UniRGB-IR to discuss their impact on model performance. As shown in Table~\ref{tab:attention}, the detector achieves the best performance with linear complexity by utilizing deformable attention. Thus, the deformable attention is more suitable for our framework and utilized as the default configuration. It is worth noting that it can be replaced by other attention mechanisms to further achieve superior performance.

\begin{figure}[!tbp]
    \centering
    \includegraphics[width=0.8\linewidth]{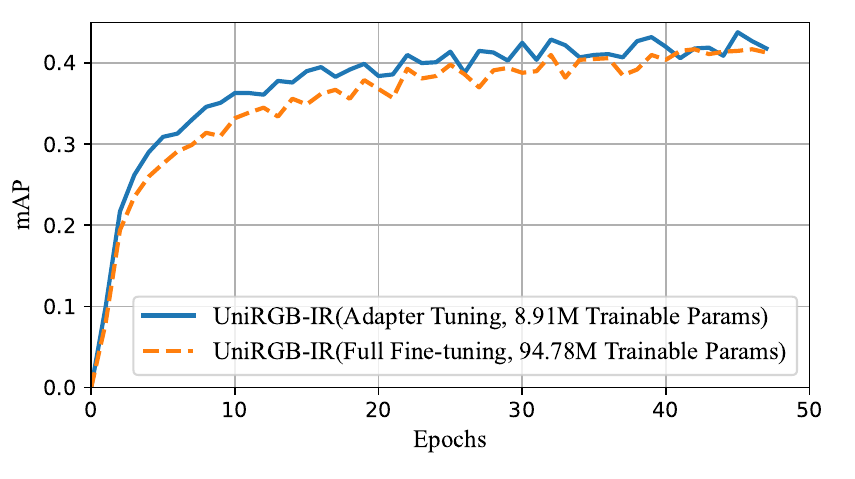}

    \caption{Training efficiency analysis on the FLIR dataset.}

    \label{fig:efficiency}
\end{figure}

\subsection{Visualization Analysis}
\noindent\textbf{Intermediate results.}
To illustrate the effectiveness of the SFI module, we visualize the intermediate results on the FLIR dataset. From $F_{mfp}$ and $F_{sfi}$ in Figure~\ref{fig:intermediate}, we can find that the foreground objects in the $F_{sfi}$ become salient through the SFI module. Furthermore, we also visualize the tSNE maps of $F_{mfp}$--$F_{vit}$ and $F_{sfi}$--$F_{vit}$ respectively. After using the SFI module, the distribution of injected features $F_{sfi}$ is more concentrated than the distribution of ViT features $F_{vit}$, indicating that the required richer RGB-IR features can be well supplemented into the ViT model through the SFI module.

\noindent\textbf{Training efficiency.}
We further plot the mAP curves for each epoch of UniRGB-IR with different training paradigm to demonstrate the efficiency of UniRGB-IR, as shown in Figure~\ref{fig:efficiency}. During the training process, all hyperparameters of the two models are the same. From Figure~\ref{fig:efficiency}, we can find that the convergence speed of the adapter tuning paradigm surpasses that of the full fine-tuning strategy. Moreover, by utilizing the adapter tuning paradigm, our UniRGB-IR achieves superior performance with a smaller number of trainable parameters (about 10\% of the full fine-tuning model). The above results verify the efficiency of our method.

\section{Conclusion}
In this paper, we proposed an efficient and scalable framework (named UniRGB-IR) for RGB-IR semantic tasks. The framework contains a Multi-modal Feature Pool module and a Supplementary Feature Injector module. The former extracts contextual multi-scale features from two modality images, and the latter adaptively injects the features into the transformer model. These two modules can be efficiently optimized to complement the pre-trained foundation model with richer RGB-IR features.
To evaluate the effectiveness of our method, we incorporated the ViT-Base model into the framework as the pre-trained foundation model and performed various RGB-IR semantic tasks. Extensive experiments verify that our UniRGB-IR can be effectively leveraged as a unified framework for RGB-IR downstream tasks. We believe that our method can be applied to more multi-modal real-world applications.

\begin{acks}
This work was supported by the Fundamental Research Funds for the Central Universities.

\end{acks}
\bibliographystyle{ACM-Reference-Format}
\balance
\bibliography{sample-base}

\end{document}